\documentclass{bmvc2k}

\usepackage{algorithm}
\usepackage{algpseudocode}
\usepackage{amsmath}
\usepackage{lipsum}

\usepackage{algorithm}
\usepackage{graphbox}
\usepackage{subfigure}
\usepackage{color}
\usepackage{setspace}
\usepackage{subfigure}
\usepackage{multirow}
\usepackage{float}
\usepackage{epsfig}
\usepackage{graphicx}
\usepackage{amsmath}
\usepackage{amssymb}
\usepackage{algorithm}
\usepackage{algpseudocode}
\usepackage{threeparttable}
\usepackage{array}
\usepackage{booktabs}
\usepackage{multirow}

\title{A New Perspective to Boost Vision Transformer for Medical Image Classification}

\addauthor{Yuexiang Li}{vicyxli@tencent.com}{1}
\addauthor{Yawen Huang}{yawenhuang@tencent.com}{1}
\addauthor{Nanjun He}{nanjunhe@tencent.com}{1}
\addauthor{Kai Ma}{kylekma@tencent.com}{1}
\addauthor{Yefeng Zheng}{yefengzheng@tencent.com}{1}

\addinstitution{
 Tencent Jarvis Lab\\
 Shenzhen\\
 China
}

\runninghead{Student, Prof, Collaborator}{BMVC Author Guidelines}

\begin{document}

\maketitle

\begin{abstract}
   Transformer has achieved impressive successes for various computer vision tasks. However, most of existing studies require to pretrain the Transformer backbone on a large-scale labeled dataset (\emph{e.g.,} ImageNet) for achieving satisfactory performance, which is usually unavailable for medical images. Additionally, due to the gap between medical and natural images, the improvement generated by the ImageNet pretrained weights significantly degrades while transferring the weights to medical image processing tasks. In this paper, we propose Bootstrap Own Latent of Transformer (BOLT), a self-supervised learning approach specifically for medical image classification with the Transformer backbone. Our BOLT consists of two networks, namely online and target branches, for self-supervised representation learning. Concretely, the online network is trained to predict the target network representation of the same patch embedding tokens with a different perturbation. To maximally excavate the impact of Transformer from limited medical data, we propose an auxiliary difficulty ranking task. The Transformer is enforced to identify which branch (\emph{i.e.,} online/target) is processing the more difficult perturbed tokens. Overall, the Transformer endeavours itself to distill the transformation-invariant features from the perturbed tokens to simultaneously achieve difficulty measurement and maintain the consistency of self-supervised representations. The proposed BOLT is evaluated on three medical image processing tasks, \emph{i.e.,} skin lesion classification, knee fatigue fracture grading and diabetic retinopathy grading. The experimental results validate the superiority of our BOLT for medical image classification, compared to ImageNet pretrained weights and state-of-the-art self-supervised learning approaches.
\end{abstract}

\section{Introduction}
\label{sec:intro}
Recently, vision Transformer (ViT) \cite{dosovitskiy2020} and its variants \cite{YuanLi_2021,WenhaiWang_2021,LiuZe_2021} has been introduced for various computer vision tasks (\emph{e.g.,} image classification \cite{dosovitskiy2020,Lanchantin_2021_CVPR}, object detection \cite{zhu2020deformable,Dai_2021_CVPR_transformer}, semantic segmentation \cite{Zheng_2021_CVPR,Wang_2021_CVPR_transformer} and medical image processing \cite{JeyaMaria_2021,Yuanfeng_2021,Yunhe_2021,GePeng_2021,Yinglin_2021}) and gained increasing attentions from the community. The common ViT usually requires pretrainig on large-scale natural image datasets, \emph{e.g.,} ImageNet, to achieve the satisfactory performance. For natural images, the labels for pretraining dataset can be efficiently obtained by crowdsourcing, as even ordinary people possess the ability to effectively identify and label objects in natural images. However, the same strategy cannot be adopted for medical images, as professional expertise is mandatory for high-quality medical image annotations. Hence, the limited amount of annotated medical data is the major obstacle for the improvement of diagnosis accuracy even with the powerful vision Transformer.

Self-supervised learning (SSL) approach is a potential solution to tackle the challenge of insufficient annotated data. The typical self-supervised learning formulates a proxy task to extract representative features from unlabeled data, which can boost the accuracy of subsequent target task. Existing studies have proposed various proxy tasks, including grayscale image colorization \cite{larsson_colorization_2017}, patch re-ordering \cite{NorooziVFP18}, and context restoration \cite{pathakCVPR}. The SSL was firstly brought to medical image processing by Zhang \emph{et al.} \cite{Zhang_2017}. Concretely, the neural network was pretrained with a proxy task that sorted the 2D slices from the conventional 3D medical volumes for the subsequent fine-grained body part recognition. Zhu \emph{et al.} \cite{ZHU2020101746} enforced 3D networks to play a Rubik's cube game for pretraining, which can be seen as an extension of 2D Jigsaw puzzles \cite{Noroozi_2016_ECCV}. Contrastive learning \cite{1640964} has been recently popularized for self-supervised representation learning. These approaches enforce neural networks to spontaneously exploit useful information from pairs of positive and negative samples, instead of permuting the contextual information of images for self-supervised signal formulation. He \emph{et al.} \cite{He_2020_CVPR_MOCO} firstly introduced the idea of contrastive learning into the area of self-supervised learning. They proposed an approach, namely MoCo, which addressed the problem of large number of negative samples for contrastive learning by maintaining a memory bank of negative samples. Following the direction, various contrastive-learning-based self-supervised approaches have been proposed \cite{pmlr-v119-chen20j, Xinlei_2020, Grill2020, ChenX_2021, Pan_2021_CVPR, Wang_2021_CVPR, ZhendaXie_2021}.
Inspired by the success of self-supervised learning for CNNs, researchers began to make their efforts to ViT. Atito \emph{et al.} \cite{AtitoS_2021} directly utilized the existing SSL approaches, including rotation prediction, contrastive learning and image restoration, to pretrain vision Transformers. Several studies \cite{TMI_ref_1, TMI_ref_2} have been proposed along this direction. {\itshape However, taking the architecture difference between CNN and ViT into account, i.e., CNN takes the whole image as input, while the input of ViT is the embedding tokens of image tiles, the self-supervised learning approach specifically for ViT is worthwhile to develop.}

In the recent study, Chen \emph{et al.} \cite{ChenX_2021} proposed MoCo V3 as a token-based constrastive learning approach, specifically for ViT to extract self-supervised features from raw data. The network pretrained with MoCo V3 outperformed the ImageNet-pretrained one, which demonstrated the effectiveness of token-based self-supervised learning. In this paper, we follow the direction and propose a token-wise perturbation based self-supervised learning framework specifically for medical image classification with vision Transformer, namely Bootstrap Own Latent of Transformer (BOLT). Similar to the existing Bootstrap Your Own Latent (BYOL) \cite{Grill2020}, our BOLT consists of two networks, namely online and target branches, for self-supervised representation learning. Instead of image-wise transformation adopted by BYOL, the online network of our BOLT is trained to predict the target network representation of the same patch embedding tokens with a different perturbation. Moreover, to encourage the vision Transformer to deeply exploit useful information from limited medical data, we propose an auxiliary difficulty ranking task. The difference between the original patch embedding tokens and the perturbed ones is measured as the difficulty (\emph{i.e.,} the larger difference means more difficult for the vision Transformer to process), which is then adopted as the supervision signal. In other words, the vision Transformer is required to identify which branch (online/target) is processing the more difficult perturbed tokens. Under the co-supervision of the two tasks, the vision Transformer is encouraged to endeavour itself to distill the transformation-invariant features from the perturbed tokens, which should be capable for simultaneous difficulty measurement and maintain the consistency of self-supervised representations. In summary, the main contributions of our work can be concluded into four-fold:
\begin{itemize}
  \item [$\bullet$] A token perturbation based self-supervised learning approach, namely BOLT, specifically designed for vision Transformer is proposed. A token perturbation module is integrated to the existing BYOL framework for the more effective ViT pretraining.

  \item [$\bullet$] An auxiliary self-supervised task, \emph{i.e.,} difficulty ranking, is proposed to encourage ViTs to deeply exploit useful information from limited medical data. The self-supervised signal of this auxiliary task also derives from the perturbed tokens generated by our perturbation module.  To our best knowledge, this is the first SSL framework based on the difficulty-awareness paradigm.

  \item [$\bullet$] The proposed BOLT is evaluated on three medical image processing tasks, \emph{i.e.,} skin lesion classification, knee fatigue fracture grading and diabetic retinopathy grading. The experimental results demonstrate the superiority of our BOLT, compared to the widely-used ImageNet pretrained weights.

  \item [$\bullet$] Last but not least, we pretrain the ViT using different self-supervised learning approaches on a large-scale private fundus image dataset captured from a collaborating hospital for diabetic retinopathy grading task. The dataset consists of 350,000 fundus images of normal cohort and patients with various diseases, which may be the largest fundus image dataset in the worldwide. The pretraining on our private large-scale dataset is verified to benefit the related downstream target task. To advance the development of automated fundus image processing, we will release the ViT pretrained models to the community.
\end{itemize}

\begin{figure*}[!tb]
  \begin{center}
    \includegraphics[width=\linewidth]{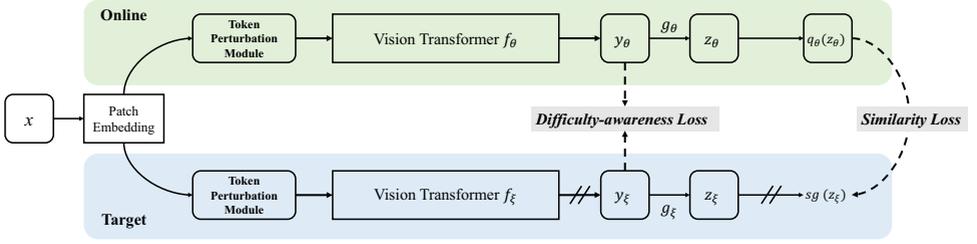}
  \end{center}\caption{The architecture of our BOLT framework. Compared to the original BYOL, our BOLT consists of two main revisions: 1) The proposed BOLT generates two views of embedding tokens for self-supervised learning; 2) A novel difficulty-awareness loss is proposed to encourage the ViT to deeply exploit useful information from raw data. $sg(.)$ means stop-gradient.}\label{fig:BOLT_pipeline}
  \vspace{-3mm}
\end{figure*}

\section{Method}
In this section, we introduce the proposed BOLT framework in details. The pipeline of our Bootstrap Own Latent of Transformer (BOLT) is illustrated in Fig.~\ref{fig:BOLT_pipeline}. Similar to BYOL, the proposed BOLT adopts two branches to extract useful information from raw data, \emph{i.e.,} the online and target branches. The online branch consists of a set of weights $\theta$, including a vision Transformer $f_{\theta}$, a projector $g_{\theta}$ and a predictor $q_{\theta}$. The target branch is of the same architecture with a different set of weights $\xi$. The target branch generates the regression targets for the online branch to learn, and its parameters $\xi$ are an exponential moving average of the online branch parameters $\theta$, which can be defined as:
\begin{equation}
  \xi \leftarrow \tau \xi+(1-\tau) \theta \label{eq:EMA}
\end{equation}
where $\tau \in [0, 1]$ is the decay rate.

\begin{figure}[!tb]
  \begin{center}
    \includegraphics[width=0.5\linewidth]{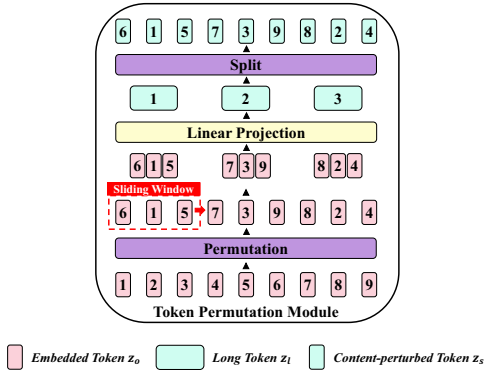}
  \end{center}\caption{The architecture of the proposed token perturbation module. The module consists of three operations (\emph{i.e.,} permutation, linear projection and split) to perturb the order and content of embedded tokens. Note that nine embedding tokens in this figure are taken as an example. The exact number ($N$) of embedding tokens is decided by $\frac{HW}{P^2}$, where $H$ and $W$ are the height and width of the original image, respectively, and ($P$, $P$) is the size of each image patch.}\label{TPM}
  \vspace{-3mm}
\end{figure}

Compared to the existing BYOL \cite{Grill2020}, the proposed BOLT has two differences: First, instead of image-based perturbation, we implement a token-based perturbation module for the constrastive learning. The underlying reason for the token-based perturbation is that the vision Transformer is insensitive to the order of input embedded tokens due to the mechanism of self-attention, which neutralizes the effectiveness of typical image-based transformation (\emph{e.g.,} Jigsaw puzzle permutation \cite{Noroozi_2016_ECCV}) made to the self-supervised learning of ViT. Inspired by recent studies \cite{YuanLi_2021,Chu_2021}, our token perturbation module involves permutation, fusion and split operations to simultaneously disarrange the order and content of tokens. Second, since the recent study \cite{9433822} demonstrated the difficulty-awareness can boost the performance of CNNs, a difficulty-awareness auxiliary task, \emph{i.e.,} requiring the ViT to identify which branch (online/target) is processing the more difficult perturbed tokens, is integrated to the existing BYOL framework.

\subsection{Token Perturbation Module}
Instead of permuting the image content, we propose a token perturbation module to perturb the order and content of embedded tokens for the self-supervised learning of a vision Transformer. The architecture of our token perturbation module is presented in Fig.~\ref{TPM}, which involves three operations, \emph{i.e.,} permutation, linear projection and split.

\vspace{2mm}
\noindent {\bf Permutation.} 
Similar to the typical vision Transformer, the input image $x\in\mathbb{R}^{H\times W\times C}$ is cropped into a sequence of flattened 2D patches $x_p\in \mathbb{R}^{N\times (P^2C)}$, where $H$ and $W$ are the height and width of the original image, respectively, $C$ is the number of channels, ($P$, $P$) is the size of each image patch, and $N=\frac{HW}{P^2}$ is the resulting number of patches. Therefore, the embedded tokens $z_o$ can be written as:
\begin{equation}
  z_o = [x^1_p\mathbf{E};x^2_p\mathbf{E};\cdots;x^N_p\mathbf{E}],
\end{equation}
where $\mathbf{E}\in\mathbb{R}^{(P^2C)\times D}$ is a trainable linear projection ($D$ is the latent vector size of the vision Transformer). Then, the permuted tokens $z_p$ are obtained using a permutation operation ($Perm(\cdot)$), which randomly disarranges the order of $z_o$: $z_p = Perm(z_o)$. Fig.~\ref{TPM} shows an example, the order of $z_o$ is disarranged to $[z_o^6;z_o^1;z_o^5;z_o^7;z_o^3;z_o^9;z_o^8;z_o^2;z_o^4]$.

\vspace{2mm}
\noindent {\bf Linear Projection.} 
After the permutation, we concatenate $M$ adjacent tokens using a sliding window with a stride $S=\frac{W}{P}$, which results in $K=\frac{N}{S}$ long tokens ($z_p'$) with the length of $M\times D$. The obtained tokens are then fed to a linear projection layer ($\mathbf{E}_{fuse}\in \mathbb{R}^{MD \times SD}$) for information fusion, which yields $K$ content-perturbed long tokens ($z_l$):
\begin{equation}
  z_l = z_p'\mathbf{E}_{fuse}.
\end{equation}


\vspace{2mm}
\noindent {\bf Split.} 
As previously mentioned, the typical vision Transformer uses the constant latent vector size $D$ through all of its layers; hence, the fused tokens with the length of $S\times D$ need to be reshaped back to the length of $D$ to fulfill the input requirement of ViT. To achieve that, the proposed token perturbation module adopts a split operation to separate each long token into $S$ $D$-length tokens. The splitted tokens ($z_s$) is then fed to ViT for self-supervised learning.

\subsection{Loss Function}
As shown in Fig.~\ref{fig:BOLT_pipeline}, our BOLT is jointly supervised by two loss functions, \emph{i.e.,} similarity loss and difficulty-awareness loss. The similarity loss is consistent to the existing BYOL framework. Concretely, for a set of embedded tokens $z_o$, our BOLT produces two augmented perturbed tokens $z_t$ and $z'_t$ for online and target branches, respectively. The perturbed tokens $z_t$  are then fed to a ViT $f_{\theta}$, which yields a representation $y_{\theta} = f_{\theta}(z_t)$ and a projection $z_{\theta} = g_{\theta}(y_{\theta})$. For the perturbed tokens for the target branch, a representation $y_{\xi} = f_{\xi}(z'_t)$ and a projection $z_{\xi} = g_{\xi}(y_{\xi})$ are accordingly generated. Consistent to BYOL, a prediction network $q_{\theta}(.)$ is adopted to yield the prediction of $z_{\xi}$ and $l_2$-norm is calculated for network training:
\begin{equation}
  \mathcal{L}_{\theta} = \left \| q_{\theta}(z_{\theta}) - z_{\xi} \right \|_2^2
\end{equation}
where $\theta$ denotes the network weights of the online branch including $f_{\theta}$, $g_{\theta}$ and $q_{\theta}$. The loss $\mathcal{L}_{\theta}^{BOLT} = \mathcal{L}_{\theta} + \tilde{\mathcal{L}}_{\theta}$ only optimizes the weights of online branch $\theta$, where $\tilde{\mathcal{L}}_{\theta}$ is the symmetric loss of $\mathcal{L}_{\theta}$ by feeding $z'_t$ and $z_t$ to online and target branches, respectively.

\vspace{2mm}
\noindent {\bf Difficulty-awareness Loss.} Apart from the similarity loss, inspired by the curriculum learning \cite{HK2019MICCAI}, we propose an auxiliary task---identifying which branch is processing the tokens with a larger level of perturbation. Such an auxiliary task can drive ViTs to self-adaptively pay more attention on the hard case and accordingly better exploit the semantic information from the embedded tokens, since they are required to understand the content of tokens for the accurate difficulty ranking.

To formulate the auxiliary task, the self-supervised signal needs to be first generated. Assuming the perturbed tokens feeding to online and target branches as $z_t$ and $z'_t$, respectively, the self-supervised signal $y_{self}$ can be defined as:
\begin{equation}
  {\footnotesize
  y_{self} = \left\{\begin{matrix}
    0, & MSE(Perm^{-1}_{z_t}(z_t) - z_o) < MSE(Perm^{-1}_{z'_t}(z'_t) - z_o)         \\
    1, & MSE(Perm^{-1}_{z_t}(z_t) - z_o) \geqslant MSE(Perm^{-1}_{z'_t}(z'_t) - z_o)
  \end{matrix}\right.}
\end{equation}
where $MSE(.)$ is the mean squared error function; $Perm^{-1}(.)$ is the inverse permutation operation rearranging the perturbed tokens back to the original order. 

After the self-supervision is obtained, the features extracted by the online and target ViTs (\emph{i.e.,} $y_{\theta}$ and $y_{\xi}$) are concatenated ($Cat(.)$) and sent to a fully-connected layer ($FC(.)$) for difficulty classification. Specifically, the process can be written as:
\begin{equation}
  \mathcal{L}_{f_{\theta}}^{Diff} = -y_{self}*log(p) - (1-y_{self})*log(1-p))
\end{equation}
where $p=FC(Cat(y_{\theta},y_{\xi})))$ is the probability of $y_{self}=1$. Similar to $\mathcal{L}_{\theta, \xi}^{BOLT}$, the difficulty-awareness loss only optimizes the online branch ($f_\theta$).

We notice that the recent study \cite{9433822} has already proposed a difficulty-awareness loss for scleral spur localization. Hence, it is worthwhile to emphasize the difference between it and our loss function. Concretely, Tao \emph{et al.} \cite{9433822} explicitly enforced networks to predict the Dice score of input images using segmentation ground truth to achieve difficulty-awareness. Due to the lack of manual annotations, few study introduces the idea of difficulty-awareness for self-supervised learning (SSL). In this study, we obtain the difficulty-related information in a self-supervised manner using the token perturbation module, and implicitly formulate the difficulty-ranking proxy task. To our best knowledge, this is the first SSL framework based on the difficulty-awareness paradigm.

\vspace{2mm}
\noindent {\bf Overall Objective.} Combining the aforementioned loss functions ($\mathcal{L}^{BOLT}$ and $\mathcal{L}^{Diff}$), the full objective $\mathcal{L}$ for the optimization of the online branch can be written as:
\begin{equation}
  \mathcal{L} = \mathcal{L}^{BOLT}_{\theta}+ \alpha \mathcal{L}^{Diff}_{f_{\theta}}
\end{equation}
where $\alpha=0.1$ is the loss weight of $\mathcal{L}_{f_{\theta}}^{Diff}$. According to Eq.~(\ref{eq:EMA}), the weights of target branch $\xi$ are updated via exponential moving average.

\section{Experiments}
We evaluate the proposed BOLT on three target tasks, including skin lesion classification, knee fatigue grading and diabetic retinopathy grading, using publicly available and private datasets.
Conventional self-supervised learning approaches often pretrain the models on a large-scale unlabeled dataset (\emph{i.e.,} proxy set), and then finetune them on the relatively smaller target set. In this paper, three different medical image processing tasks are involved for performance evaluation and the corresponding proxy and target datasets (example images are shown in {\itshape Supplementary Material}) for each task are introduced in the followings:

\vspace{2mm}
\noindent {\bf Skin Lesion Classification.} 
The publicly available ISIC 2019 dataset\footnote{https://challenge2019.isic-archive.com/} is used to validate the effectiveness of the proposed BOLT. Specifically, the dataset \cite{tschandl2018ham10000} is provided by the ISIC 2019 challenge, which encourages researchers to develop the automated systems predicting eight skin disease categories with dermoscopic images, \emph{i.e.,} squamous cell carcinoma, melanocytic nevus, benign keratosis, actinic keratosis, dermatofibroma, basal cell carcinoma, vascular lesion, and melanoma. The whole ISIC 2019 dataset, consisting of over 20,000 dermoscopic images, is adopted as the \textbf{\itshape proxy set}. Due to the class imbalance problem of original ISIC dataset, consistent to \cite{Continual_Dual_Distillation}, 628 images are randomly sampled from each class to establish a balanced \textbf{\itshape target set}. It is worthwhile to mention that the images from the two classes consisting of fewer than 628 images are all taken into the target set.
After that, the balanced target set with 4,260 images is randomly separated into training, validation and test sets based on the ratio of 70:10:20. Note that the ViT is first pretrained on the proxy set and finetuned on the training and validation sets, and then evaluated on the test set.

\vspace{2mm}
\noindent {\bf Knee Fatigue Grading.} 
The publicly available MURA dataset\footnote{https://stanfordmlgroup.github.io/competitions/mura/} (musculoskeletal radiographs) \cite{Rajpurkar_MIDL_2018}, which is a large dataset of bone X-rays (over 40,000 images), is adopted as the \textbf{\itshape proxy set} to pretrain ViTs for the subsequent target task (\emph{i.e.,} knee fatigue grading). For the knee fatigue grading, 2,725 X-ray images are collected from a collaborating hospital as the \textbf{\itshape target set} \cite{anonymous}. The positions of fatigue fracture are different, \emph{i.e.,} navicular bone, tibia and fibula. Each X-ray image is labeled by three physicians, and the final grade is decided via majority-voting. In particular, the target set has 1,785 normal, 190 grade-1, 452 grade-2, 196 grade-3 and 102 grade-4 cases, respectively.
For the evaluation on our private knee fatigue grading dataset, the target set is divided to training, validation and test sets according to the ratio of 70:10:20. Similar to \cite{anonymous}, due to the imbalance problem (normal \emph{vs.} fatigue fracture and grade-2 \emph{vs.} other fracture grades), an equal number (20) of test images from each category are randomly sampled to form an uniform-distribution set for performance evaluation, instead of using the whole test set.


\vspace{2mm}
\noindent {\bf Diabetic Retinopathy Grading.} 
For the diabetic retinopathy grading task, we pretrain the ViT on a large-scale private dataset captured from a collaborating hospital (\textbf{\itshape proxy set}), with approval obtained from the institutional review board of the hospital. The dataset consists of 350,000 fundus images of normal cohort and patients with various diseases. Then, the pretrained ViT is finetuned on the publicly available APTOS 2019 blindness detection dataset (\textbf{\itshape target set}) for performance evaluation.\footnote{https://www.kaggle.com/c/aptos2019-blindness-detection} In particular, there are 3,662 fundus images contained in the target set and the severity of diabetic retinopathy (DR) can be classified to four grades,  
\emph{i.e.,} normal (1,805), mild DR (370), moderate DR (999), severe DR (193) and proliferative DR (295). Consistent to \cite{Shaoteng_MICCAI}, a five fold cross-validation is conducted on this dataset.\footnote{The ViT pretrained on our private large-scale dataset may benefit the related downstream target tasks. To advance the development of automated fundus image processing, we will release the ViT pretrained models to the community soon.} 

\vspace{2mm}
\noindent {\bf Baselines \& Evaluation Criterion.}
To demonstrate the effectiveness of our BOLT pretraining, we finetune ViTs with ImageNet pretrained weights on the target tasks and evaluate their performances on the test set. Consistent to MoCo V3 \cite{ChenX_2021}, the basic ViT-B/16 is adopted as backbone. The original BYOL \cite{Grill2020}, state-of-the-art self-supervised learning approach SimSam \cite{Chen_2021_CVPR} and token-based self-supervised learning approach MoCo V3 \cite{ChenX_2021} are assessed for comparison. It is worthwhile to mention that the backbones of representation networks of BYOL and SimSam implemented in this study are ViT-B/16. The average classification accuracy (ACC) is adopted as metric for the performance evaluation.

\begin{table}[!tb]
  \small
  \caption{The classification accuracy (ACC) presented in percentage (\%) of ViTs using different training strategies with different amounts of training data on the ISIC 2019 test set.}\label{tab:ISIC2019}
  \begin{center}
    \begin{tabular}{l|c|c|c}
      \hline
      {}                              & \bf 100\% & \bf 50\% & \bf 10\% \\\hline\hline
      Train-from-scratch              & 39.4      & 35.2     & 31.3     \\\hline
      ImageNet Pretrained             & 80.5      & 76.1     & 62.1     \\\hline
      SimSam \cite{Chen_2021_CVPR}    & 79.9      & 75.9     & 61.2     \\\hline
      BYOL \cite{Grill2020}           & 80.1      & 75.4     & 61.3     \\\hline
      MoCo V3 \cite{ChenX_2021}       & 80.3      & 75.2     & 61.2     \\\hline
      BOLT w./o. $\mathcal{L}^{Diff}$ & 80.8      & 75.8     & 62.1     \\\hline
      BOLT (ours)                     & \bf 81.5  & \bf 76.6 & \bf 62.4 \\\hline\hline
      ImageNet Pretrained ResNet-50   & 75.7      & 72.5     & 61.2     \\\hline
    \end{tabular}
  \end{center}   
  \vspace{-6mm}
\end{table}

\subsection{Performance Evaluation}
In this section, we evaluate the effectiveness of different training strategies on different datasets and present the experimental results. The widely-used ImageNet pretrained ResNet-50 is also adopted as a baseline for comparison. Some detailed discussions are presented in {\itshape Supplementary Material}.

\vspace{2mm}
\noindent {\bf Skin Lesion Classification.}
First, the different training strategies are evaluated on the publicly available ISIC 2019 dataset. The evaluation results of models finetuned with all training data ($100\%$) on the test set are listed in Table~\ref{tab:ISIC2019}. The ImageNet pretrained ViT is observed to surpass the ImageNet pretrained ResNet-50 by a large margin (\emph{i.e.,} $+4.8\%$), which demonstrates the superiority of ViT for medical image classification. Compared to the state-of-the-art self-supervised learning approaches (\emph{i.e.,} SimSam, BYOL and MoCo V3), our token-based BOLT achieves a higher ACC ($80.8\%$). By using the difficulty-awareness loss ($\mathcal{L}^{Diff}$), the ACC of BOLT can be further improved to $81.5\%$, which outperforms the runner-up (MoCo V3) by a margin of $+1.2\%$.

The goal of self-supervised learning approach primarily is to deal with the insufficient training data. Hence, to better verify the superiority of our BOLT approach, we conduct an experiment to assess the performance of BOLT pretrained ViTs with different numbers of labeled samples used for finetuning (\emph{i.e.,} $10\%$ and $50\%$ in Table~\ref{tab:ISIC2019}). It can be observed that our BOLT can effectively tackle the situation with few labeled training samples---the proposed BOLT with difficulty-awareness loss achieves the best ACC under both $50\%$ and $10\%$ settings.



\vspace{2mm}
\noindent {\bf Knee Fatigue Grading.}
Consistent to the previous study \cite{anonymous}, apart from classification accuracy, the F1 score is also adopted for performance evaluation. The experimental results on the uniform test set are listed in Table~\ref{tab:grading}.
As shown, the ViT pretrained with the proposed BOLT outperforms the ones using existing self-supervised learning approaches and the ImageNet pretrained weights, \emph{i.e.,} an ACC of $54.0\%$ is achieved ($+2.0\%$ higher than the runner-up). Similar trend to ISIC 2019 is observed---the ACC of ImageNet pretrained ViT ($51\%$) is significantly higher than that of ImageNet pretrained ResNet-50 ($36\%$), demonstrating the effectiveness of ViT backbone. We notice that the improvements to train-from-scratch yielded by pretraining are more obvious on our knee fatigue grading dataset (over $+20\%$), compared to the skin lesion classification task. The reason may be that the target set of knee fatigue grading contains less training samples (around 1,000 X-ray images); thus, it is more difficult to well train the model from scratch, compared to the skin lesion classification task with a target set of 4,260 images.

\begin{table*}[!tb]
  \begin{center}
  \caption{The accuracy (ACC and F1 score) presented in percentage (\%) of different training strategies on knee fatigue grading and diabetic retinopathy grading tasks.}\label{tab:grading}
  \footnotesize
    \begin{tabular}{l|p{1.5cm}<{\centering} p{1.5cm}<{\centering}|p{2cm}<{\centering} p{2cm}<{\centering}}
      \hline
      \multirow{2}{*}{}               & \multicolumn{2}{c|}{\bf Knee Fatigue Grading} & \multicolumn{2}{c}{\bf Diabetic Retinopathy Grading}                       \\
      \cline{2-5}
      {}                              & \bf ACC                                       & \bf F1                                               & \bf ACC  & \bf F1   \\\hline\hline
      Train-from-scratch              & 30.0                                          & 23.1                                                 & 71.0     & 65.3     \\\hline
      ImageNet Pretrained             & 51.0                                          & 49.4                                                 & 83.6     & 83.2     \\\hline
      SimSam \cite{Chen_2021_CVPR}    & 52.0                                          & 51.1                                                 & 84.5     & 84.3     \\\hline
      BYOL \cite{Grill2020}           & 51.0                                          & 50.2                                                 & 84.8     & 84.7     \\\hline
      MoCo V3 \cite{ChenX_2021}       & 52.0                                          & 51.2                                                 & 84.7     & 84.3     \\\hline
      BOLT w./o. $\mathcal{L}^{Diff}$ & 52.0                                          & 51.2                                                 & 85.4     & 85.3     \\\hline
      BOLT (ours)                     & \bf 54.0                                      & \bf 53.6                                             & \bf 85.9 & \bf 85.8 \\\hline\hline
      ImageNet Pretrained ResNet-50   & 36.0                                          & 31.7                                                 & 81.7     & 82.0     \\\hline
    \end{tabular}
    \end{center}
    \vspace{-6mm}
\end{table*}

\vspace{2mm}
\noindent {\bf Diabetic Retinopathy Grading.} Consistent to \cite{Shaoteng_MICCAI}, we split the APTOS 2019 dataset into five folds for cross-validation and adopt the F1 score for performance evaluation. The grading accuracy of models using different training strategies is shown in Table~\ref{tab:grading}. The proposed BOLT pretrained ViT achieves the best ACC ($85.9\%$) and F1 score ($85.8\%$) among the listed approaches, which are $+1.1\%$ and $+1.1\%$ higher than the original BYOL, respectively.

\section{Conclusion}
In this paper, a self-supervised learning approach, termed Boostrap Own Latent of Transformer (BOLT), was proposed specifically for medical image classification with the vision Transformer backbone. The proposed BOLT involved online and target branches, which extracted the self-supervised representation from raw data via contrastive learning. Concretely, the online network was trained to predict the target network representation of the same patch embedding tokens with a different perturbation. 
Furthermore, we proposed an auxiliary difficulty ranking task to enable the vision Transformer to exploit diverse information from the limited medical data. The difference between the original patch embedding tokens and the perturbed ones was calculated as the difficulty measurement (\emph{i.e.,} the larger difference means more difficult for the vision Transformer to process), which was then adopted as the supervision signal for self-supervised learning. 
The vision Transformer was trained to identify the branch (online/target) processing for the more difficult perturbed tokens, which enabled it to distill the transformation-invariant features from the perturbed tokens. 
The experimental results on three medical image classification tasks (\emph{i.e.,} skin lesion classification, knee fatigue fracture grading and dabetic retinopathy grading) demonstrated the effectiveness of the proposed BOLT.
We notice several limitations of this study and plan to address them in the future works:

\vspace{2mm}
\noindent {\bf Extension to Medical Image Segmentation Task.} The proposed BOLT can be easily extended to medical image segmentation in a similar way like \cite{ZHU2020101746}, \emph{i.e.,} pretraining the encoder and using a random initialization for the decoder. Yet, the randomly initialized decoder may neutralize the performance improvement. Therefore, we plan to explore a more effective way extending our pretrained ViTs for medical image segmentation task in the future.

\vspace{2mm}
\noindent {\bf Pretrained Weights for ViT Variants.} Recently, many powerful ViT-based backbones, such as Swin Transformer \cite{LiuZe_2021}, have been proposed. The weights of these ViT variants pretrained on our large-scale fundus image dataset will be continuously provided in the future.

\bibliography{reference_add}

\begin{thebibliography}{41}
\providecommand{\natexlab}[1]{#1}
\providecommand{\url}[1]{\texttt{#1}}
\expandafter\ifx\csname urlstyle\endcsname\relax
  \providecommand{\doi}[1]{doi: #1}\else
  \providecommand{\doi}{doi: \begingroup \urlstyle{rm}\Url}\fi

\bibitem[Atito et~al.(2021)Atito, Awais, and Kittler]{AtitoS_2021}
Sara Atito, Muhammad Awais, and Josef Kittler.
\newblock {SiT}: {S}elf-supervised vision {T}ransformer.
\newblock \emph{arXiv preprint arXiv:2104.03602}, 2021.

\bibitem[Bao et~al.(2021)Bao, Dong, and Wei]{TMI_ref_2}
H.~Bao, L.~Dong, and F.~Wei.
\newblock {BEiT}: {BERT} pre-training of image {Transformers}.
\newblock \emph{arXiv preprint arXiv:2106.08254}, 2021.

\bibitem[Caron et~al.(2021)Caron, Touvron, Misra, Jegou, Mairal, Bojanowski,
  and Joulin]{TMI_ref_1}
M.~Caron, H.~Touvron, I.~Misra, H.~Jegou, J.~Mairal, P.~Bojanowski, and
  A.~Joulin.
\newblock Emerging properties in self-supervised vision {Transformers}.
\newblock \emph{arXiv preprint arXiv:2104.14294}, 2021.

\bibitem[Chen et~al.(2020{\natexlab{a}})Chen, Kornblith, Norouzi, and
  Hinton]{pmlr-v119-chen20j}
Ting Chen, Simon Kornblith, Mohammad Norouzi, and Geoffrey Hinton.
\newblock A simple framework for contrastive learning of visual
  representations.
\newblock In \emph{International Conference on Machine Learning},
  2020{\natexlab{a}}.

\bibitem[Chen and He(2021)]{Chen_2021_CVPR}
Xinlei Chen and Kaiming He.
\newblock Exploring simple {S}iamese representation learning.
\newblock In \emph{IEEE Conference on Computer Vision and Pattern Recognition},
  June 2021.

\bibitem[Chen et~al.(2020{\natexlab{b}})Chen, Fan, Girshick, and
  He]{Xinlei_2020}
Xinlei Chen, Haoqi Fan, Ross Girshick, and Kaiming He.
\newblock Improved baselines with momentum contrastive learning.
\newblock \emph{arXiv preprint arXiv:2003.04297}, 2020{\natexlab{b}}.

\bibitem[Chen et~al.(2021)Chen, Xie, and He]{ChenX_2021}
Xinlei Chen, Saining Xie, and Kaiming He.
\newblock An empirical study of training self-supervised vision {T}ransformers.
\newblock \emph{arXiv preprint arXiv:2104.02057}, 2021.

\bibitem[Chu et~al.(2021)Chu, Tian, Zhang, Wang, Wei, Xia, and Shen]{Chu_2021}
Xiangxiang Chu, Zhi Tian, Bo~Zhang, Xinlong Wang, Xiaolin Wei, Huaxia Xia, and
  Chunhua Shen.
\newblock Conditional positional encodings for vision {Transformers}.
\newblock \emph{arXiv preprint arXiv:2102.10882}, 2021.

\bibitem[Dai et~al.(2021)Dai, Cai, Lin, and Chen]{Dai_2021_CVPR_transformer}
Zhigang Dai, Bolun Cai, Yugeng Lin, and Junying Chen.
\newblock {UP-DETR}: Unsupervised pre-training for object detection with
  {Transformers}.
\newblock In \emph{IEEE Conference on Computer Vision and Pattern Recognition},
  2021.

\bibitem[Dosovitskiy et~al.(2021)Dosovitskiy, Beyer, Kolesnikov, Weissenborn,
  Zhai, Unterthiner, Dehghani, Minderer, Heigold, Gelly, Uszkoreit, and
  Houlsby]{dosovitskiy2020}
Alexey Dosovitskiy, Lucas Beyer, Alexander Kolesnikov, Dirk Weissenborn,
  Xiaohua Zhai, Thomas Unterthiner, Mostafa Dehghani, Matthias Minderer, Georg
  Heigold, Sylvain Gelly, Jakob Uszkoreit, and Neil Houlsby.
\newblock An image is worth 16x16 words: {T}ransformers for image recognition
  at scale.
\newblock In \emph{International Conference on Learning Representations}, 2021.

\bibitem[Gao et~al.(2021)Gao, Zhou, and Metaxas]{Yunhe_2021}
Yunhe Gao, Mu~Zhou, and Dimitris Metaxas.
\newblock {UTNet}: A hybrid {Transformer} architecture for medical image
  segmentation.
\newblock \emph{arXiv preprint arXiv:2107.00781}, 2021.

\bibitem[Grill et~al.(2020)Grill, Strub, Altche, Tallec, Richemond,
  Buchatskaya, Doersch, Pires, Guo, Azar, Piot, Kavukcuoglu, Munos, and
  Valko]{Grill2020}
Jean-Bastien Grill, Florian Strub, Florent Altche, Corentin Tallec, Pierre~H.
  Richemond, Elena Buchatskaya, Carl Doersch, Bernardo~Avila Pires,
  Zhaohan~Daniel Guo, Mohammad~Gheshlaghi Azar, Bilal Piot, Koray Kavukcuoglu,
  Remi Munos, and Michal Valko.
\newblock Bootstrap your own latent: A new approach to self-supervised
  learning.
\newblock In \emph{Advances in Neural Information Processing Systems}, 2020.

\bibitem[Hadsell et~al.(2006)Hadsell, Chopra, and LeCun]{1640964}
R.~Hadsell, S.~Chopra, and Y.~LeCun.
\newblock Dimensionality reduction by learning an invariant mapping.
\newblock In \emph{IEEE Conference on Computer Vision and Pattern Recognition},
  2006.

\bibitem[He et~al.(2020)He, Fan, Wu, Xie, and Girshick]{He_2020_CVPR_MOCO}
Kaiming He, Haoqi Fan, Yuxin Wu, Saining Xie, and Ross Girshick.
\newblock Momentum contrast for unsupervised visual representation learning.
\newblock In \emph{IEEE Conference on Computer Vision and Pattern Recognition},
  2020.

\bibitem[Ji et~al.(2021{\natexlab{a}})Ji, Chou, Fan, Chen, Huazhu~Fu, and
  Shao]{GePeng_2021}
Ge-Peng Ji, Yu-Cheng Chou, Deng-Ping Fan, Geng Chen, Debesh~Jha Huazhu~Fu, and
  Ling Shao.
\newblock Progressively normalized self-attention network for video polyp
  segmentation.
\newblock \emph{arXiv preprint arXiv:2105.08468}, 2021{\natexlab{a}}.

\bibitem[Ji et~al.(2021{\natexlab{b}})Ji, Zhang, Wang, Li, Wu, Zhang, and
  Luo]{Yuanfeng_2021}
Yuanfeng Ji, Ruimao Zhang, Huijie Wang, Zhen Li, Lingyun Wu, Shaoting Zhang,
  and Ping Luo.
\newblock Multi-compound {Transformer} for accurate biomedical image
  segmentation.
\newblock \emph{arXiv preprint arXiv:2106.14385}, 2021{\natexlab{b}}.

\bibitem[Kervadec et~al.(2019)Kervadec, Granger, and Ayed]{HK2019MICCAI}
Hoel Kervadec, Jose~DolzÉric Granger, and Ismail~Ben Ayed.
\newblock Curriculum semi-supervised segmentation.
\newblock In \emph{International Conference on Medical Image Computing and
  Computer Assisted Intervention}, 2019.

\bibitem[Lanchantin et~al.(2021)Lanchantin, Wang, Ordonez, and
  Qi]{Lanchantin_2021_CVPR}
Jack Lanchantin, Tianlu Wang, Vicente Ordonez, and Yanjun Qi.
\newblock General multi-label image classification with {Transformers}.
\newblock In \emph{IEEE Conference on Computer Vision and Pattern Recognition},
  2021.

\bibitem[Larsson et~al.(2017)Larsson, Maire, and
  Shakhnarovich]{larsson_colorization_2017}
G.~Larsson, M.~Maire, and G.~Shakhnarovich.
\newblock Colorization as a proxy task for visual understanding.
\newblock In \emph{IEEE Conference on Computer Vision and Pattern Recognition},
  2017.

\bibitem[Li et~al.(2021)Li, Wang, Lin, Lin, Wei, Zhang, Ma, Zhang, and
  Zheng]{anonymous}
Yuexiang Li, Yanping Wang, Guang Lin, Yi~Lin, Dong Wei, Qirui Zhang, Kai Ma,
  Zhiqiang Zhang, and Yefeng Zheng.
\newblock Triplet-branch network with prior-knowledge embedding for fatigue
  fracture grading.
\newblock In \emph{International Conference on Medical Image Computing and
  Computer Assisted Intervention}, 2021.

\bibitem[Li et~al.(2020)Li, Zhong, Wang, and
  Zheng]{Continual_Dual_Distillation}
Zhuoyun Li, Changhong Zhong, Ruixuan Wang, and Wei-Shi Zheng.
\newblock Continual learning of new diseases with dual distillation and
  ensemble strategy.
\newblock In \emph{International Conference on Medical Image Computing and
  Computer Assisted Intervention}, 2020.

\bibitem[Liu et~al.(2020)Liu, Gong, Ma, and Zheng]{Shaoteng_MICCAI}
Shaoteng Liu, Lijun Gong, Kai Ma, and Yefeng Zheng.
\newblock {GREEN}: a graph residual re-ranking network for grading diabetic
  retinopathy.
\newblock In \emph{International Conference on Medical Image Computing and
  Computer Assisted Intervention}, 2020.

\bibitem[Liu et~al.(2021)Liu, Lin, Cao, Hu, Wei, Zhang, Lin, and
  Guo]{LiuZe_2021}
Ze~Liu, Yutong Lin, Yue Cao, Han Hu, Yixuan Wei, Zheng Zhang, Stephen Lin, and
  Baining Guo.
\newblock Swin {Transformer}: Hierarchical vision {Transformer} using shifted
  windows.
\newblock \emph{arXiv preprint arXiv:2103.14030}, 2021.

\bibitem[Noroozi and Favaro(2016)]{Noroozi_2016_ECCV}
M.~Noroozi and P.~Favaro.
\newblock Unsupervised learning of visual representations by solving {Jigsaw}
  puzzles.
\newblock In \emph{European Conference on Computer Vision}, 2016.

\bibitem[Noroozi et~al.(2018)Noroozi, Vinjimoor, Favaro, and
  Pirsiavash]{NorooziVFP18}
M.~Noroozi, A.~Vinjimoor, P.~Favaro, and H.~Pirsiavash.
\newblock Boosting self-supervised learning via knowledge transfer.
\newblock In \emph{IEEE Conference on Computer Vision and Pattern Recognition},
  2018.

\bibitem[Pan et~al.(2021)Pan, Song, Yang, Jiang, and Liu]{Pan_2021_CVPR}
Tian Pan, Yibing Song, Tianyu Yang, Wenhao Jiang, and Wei Liu.
\newblock {VideoMoCo}: {C}ontrastive video representation learning with
  temporally adversarial examples.
\newblock In \emph{IEEE Conference on Computer Vision and Pattern Recognition},
  2021.

\bibitem[Pathak et~al.(2016)Pathak, Kr\"ahenb\"uhl, Donahue, Darrell, and
  Efros]{pathakCVPR}
D.~Pathak, P.~Kr\"ahenb\"uhl, J.~Donahue, T.~Darrell, and A.~A. Efros.
\newblock Context encoders: Feature learning by inpainting.
\newblock In \emph{IEEE Conference on Computer Vision and Pattern Recognition},
  2016.

\bibitem[Rajpurkar et~al.(2018)Rajpurkar, Irvin, Bagul, Ding, Duan, Mehta,
  Yang, Zhu, Laird, Ball, Langlotz, Shpanskaya, Lungren, and
  Ng]{Rajpurkar_MIDL_2018}
Pranav Rajpurkar, Jeremy Irvin, Aarti Bagul, Daisy Ding, Tony Duan, Hershel
  Mehta, Brandon Yang, Kaylie Zhu, Dillon Laird, Robyn~L. Ball, Curtis
  Langlotz, Katie Shpanskaya, Matthew~P. Lungren, and Andrew~Y. Ng.
\newblock {MURA}: {L}arge dataset for abnormality detection in musculoskeletal
  radiographs.
\newblock In \emph{International Conference on Medical Imaging with Deep
  Learning}, 2018.

\bibitem[Tao et~al.(2021)Tao, Yuan, Bian, Li, Ma, Ni, and Zheng]{9433822}
Xing Tao, Chenglang Yuan, Cheng Bian, Yuexiang Li, Kai Ma, Dong Ni, and Yefeng
  Zheng.
\newblock The winner of age challenge: Going one step further from keypoint
  detection to scleral spur localization.
\newblock In \emph{IEEE International Symposium on Biomedical Imaging}, 2021.

\bibitem[Tschandl et~al.(2018)Tschandl, Rosendahl, and
  Kittler]{tschandl2018ham10000}
Philipp Tschandl, Cliff Rosendahl, and Harald Kittler.
\newblock The {HAM10000} dataset, a large collection of multi-source
  dermatoscopic images of common pigmented skin lesions.
\newblock \emph{Scientific Data}, 5\penalty0 (1):\penalty0 1--9, 2018.

\bibitem[Valanarasu et~al.(2021)Valanarasu, Oza, Hacihaliloglu, and
  Patel]{JeyaMaria_2021}
Jeya Maria~Jose Valanarasu, Poojan Oza, Ilker Hacihaliloglu, and Vishal~M.
  Patel.
\newblock Medical {Transformer}: Gated axial-attention for medical image
  segmentation.
\newblock \emph{arXiv preprint arXiv:2102.10662}, 2021.

\bibitem[Wang et~al.(2021{\natexlab{a}})Wang, Xie, Li, Fan, Song, Liang, Lu,
  Luo, and Shao]{WenhaiWang_2021}
Wenhai Wang, Enze Xie, Xiang Li, Deng-Ping Fan, Kaitao Song, Ding Liang, Tong
  Lu, Ping Luo, and Ling Shao.
\newblock Pyramid vision {Transformer}: {A} versatile backbone for dense
  prediction without convolutions.
\newblock \emph{arXiv preprint arXiv:2102.12122}, 2021{\natexlab{a}}.

\bibitem[Wang et~al.(2021{\natexlab{b}})Wang, Zhang, Shen, Kong, and
  Li]{Wang_2021_CVPR}
Xinlong Wang, Rufeng Zhang, Chunhua Shen, Tao Kong, and Lei Li.
\newblock Dense contrastive learning for self-supervised visual pre-training.
\newblock In \emph{IEEE Conference on Computer Vision and Pattern Recognition},
  2021{\natexlab{b}}.

\bibitem[Wang et~al.(2021{\natexlab{c}})Wang, Xu, Wang, Shen, Cheng, Shen, and
  Xia]{Wang_2021_CVPR_transformer}
Yuqing Wang, Zhaoliang Xu, Xinlong Wang, Chunhua Shen, Baoshan Cheng, Hao Shen,
  and Huaxia Xia.
\newblock End-to-end video instance segmentation with {Transformers}.
\newblock In \emph{IEEE Conference on Computer Vision and Pattern Recognition},
  2021{\natexlab{c}}.

\bibitem[Xie et~al.(2021)Xie, Lin, Yao, Zhang, Dai, Cao, and
  Hu]{ZhendaXie_2021}
Zhenda Xie, Yutong Lin, Zhuliang Yao, Zheng Zhang, Qi~Dai, Yue Cao, and Han Hu.
\newblock Self-supervised learning with {Swin} {Transformers}.
\newblock \emph{arXiv preprint arXiv:2105.04553}, 2021.

\bibitem[Yuan et~al.(2021)Yuan, Chen, Wang, Yu, Shi, Jiang, Tay, Feng, and
  Yan]{YuanLi_2021}
Li~Yuan, Yunpeng Chen, Tao Wang, Weihao Yu, Yujun Shi, Zihang Jiang, Francis~EH
  Tay, Jiashi Feng, and Shuicheng Yan.
\newblock {Tokens-to-Token ViT}: {T}raining vision {T}ransformers from scratch
  on {ImageNet}.
\newblock \emph{arXiv preprint arXiv:2101.11986}, 2021.

\bibitem[Zhang et~al.(2017)Zhang, Wang, and Zheng]{Zhang_2017}
P.~Zhang, F.~Wang, and Y.~Zheng.
\newblock Self supervised deep representation learning for fine-grained body
  part recognition.
\newblock In \emph{International Symposium on Biomedical Imaging}, 2017.

\bibitem[Zhang et~al.(2021)Zhang, Higashita, Fu, Xu, Zhang, Liu, Zhang, and
  Liu]{Yinglin_2021}
Yinglin Zhang, Risa Higashita, Huazhu Fu, Yanwu Xu, Yang Zhang, Haofeng Liu,
  Jian Zhang, and Jiang Liu.
\newblock A multi-branch hybrid {Transformer} network for corneal endothelial
  cell segmentation.
\newblock \emph{arXiv preprint arXiv:2106.07557}, 2021.

\bibitem[Zheng et~al.(2021)Zheng, Lu, Zhao, Zhu, Luo, Wang, Fu, Feng, Xiang,
  Torr, and Zhang]{Zheng_2021_CVPR}
Sixiao Zheng, Jiachen Lu, Hengshuang Zhao, Xiatian Zhu, Zekun Luo, Yabiao Wang,
  Yanwei Fu, Jianfeng Feng, Tao Xiang, Philip~H.S. Torr, and Li~Zhang.
\newblock Rethinking semantic segmentation from a sequence-to-sequence
  perspective with {Transformers}.
\newblock In \emph{IEEE Conference on Computer Vision and Pattern Recognition},
  2021.

\bibitem[Zhu et~al.(2020{\natexlab{a}})Zhu, Li, Hu, Ma, Zhou, and
  Zheng]{ZHU2020101746}
Jiuwen Zhu, Yuexiang Li, Yifan Hu, Kai Ma, S.~Kevin Zhou, and Yefeng Zheng.
\newblock Rubik’s cube+: A self-supervised feature learning framework for
  {3D} medical image analysis.
\newblock \emph{Medical Image Analysis}, 64:\penalty0 101746,
  2020{\natexlab{a}}.

\bibitem[Zhu et~al.(2020{\natexlab{b}})Zhu, Su, Lu, Li, Wang, and
  Dai]{zhu2020deformable}
Xizhou Zhu, Weijie Su, Lewei Lu, Bin Li, Xiaogang Wang, and Jifeng Dai.
\newblock Deformable {DETR}: {D}eformable {T}ransformers for end-to-end object
  detection.
\newblock \emph{arXiv preprint arXiv:2010.04159}, 2020{\natexlab{b}}.

\end{thebibliography}

\end{document}